\documentclass[lettersize,journal]{IEEEtran}
\usepackage{amsmath,amsfonts}

\usepackage{array}
\usepackage[font=footnotesize]{subfig}
\usepackage{textcomp}
\usepackage{stfloats}
\usepackage{url}
\usepackage{verbatim}
\usepackage[algo2e]{algorithm2e}

\usepackage{times}
\usepackage{epsfig}
\usepackage{graphicx}
\usepackage{amsmath}
\usepackage{amssymb}
\usepackage{booktabs}
\usepackage{enumitem}
\usepackage{dsfont}
\usepackage{algorithm, algorithmic}
\algsetup{linenosize=\small}
\usepackage{tablefootnote}
\usepackage[flushleft]{threeparttable} 
\usepackage{balance}
\usepackage{xcolor}
\usepackage{color,colortbl}
\usepackage{overpic}
\usepackage{wrapfig}
\usepackage[normalem]{ulem}
\usepackage{amsmath,esvect}
\usepackage{orcidlink}

\usepackage[accsupp]{axessibility}

\usepackage{mathtools}
\usepackage{mathrsfs}

\usepackage{lipsum}
\usepackage{caption}
\usepackage{hyperref}

\definecolor{tabblue}{HTML}{00b4d8}

\input{insbox.tex}
\newcommand{\name}{STAR}

\usepackage{wasysym}

\usepackage{multirow}
\usepackage{multicol}
\usepackage{soul}
\setul{0.5ex}{0.3ex}

\newcommand{\myparagraph}[1]{\vspace{2mm}\noindent\textbf{#1}}

\definecolor{col_nose}{RGB}{181,228,140}
\definecolor{col_mouth}{RGB}{255,183,3}
\definecolor{col_forehead}{RGB}{189,178,255}
\definecolor{col_cheek}{RGB}{144,224,239}
\definecolor{col_table}{RGB}{175,227,246} 

\definecolor{mycolor1}{rgb}{0.85000,0.32500,0.09800}%
\definecolor{mycolor2}{rgb}{0.92900,0.69400,0.12500}%
\definecolor{mycolor3}{rgb}{0.49400,0.18400,0.55600}%
\definecolor{mycolor4}{rgb}{0.87843,0.76471,0.98824}%
\definecolor{mycolor5}{rgb}{0.46600,0.67400,0.18800}%
\definecolor{mycolor6}{rgb}{0.30100,0.74500,0.93300}%
\definecolor{mycolor7}{rgb}{0.00000,0.44700,0.74100}%

\definecolor{best_two}{RGB}{72,149,239} 
\definecolor{best}{RGB}{179,11,0} 

\usepackage{xspace}

\makeatletter
\DeclareRobustCommand\onedot{\futurelet\@let@token\@onedot}
\def\@onedot{\ifx\@let@token.\else.\null\fi\xspace}

\def\eg{\emph{e.g}\onedot} 
\def\ie{\emph{i.e}\onedot} 
 
 \def\vs{\emph{vs}\onedot}
\def\wrt{w.r.t\onedot} 

\def\iff{\emph{iff}}
\def\vs{\emph{v.s}\onedot}

\makeatother

\def\BibTeX{{\rm B\kern-.05em{\sc i\kern-.025em b}\kern-.08em
    T\kern-.1667em\lower.7ex\hbox{E}\kern-.125emX}}
\usepackage{balance}

\begin{document}

\title{{\name}: Skeleton-aware Text-based 4D Avatar Generation with In-Network Motion Retargeting}
\author{Zenghao Chai\orcidlink{0000-0003-3709-4947}, 
        Chen Tang\orcidlink{0000-0002-0108-6729}, 
        Yongkang Wong\orcidlink{0000-0002-1239-4428},~\IEEEmembership{Member,~IEEE}, and 
        Mohan Kankanhalli\orcidlink{0000-0002-4846-2015},~\IEEEmembership{Fellow,~IEEE}
\thanks{Z. Chai, Y. Wong, and M. Kankanhalli are with the School of Computing, National University of Singapore, Singapore, 117417. Email: zenghaochai@gmail.com, yongkang.wong@nus.edu.sg, mohan@comp.nus.edu.sg. C. Tang is with the Department of Computer Science and Technology, Tsinghua University, China, 100084. Email: tangc@tsinghua.edu.cn.}
\thanks{Mohan Kankanhalli is the corresponding Author.}}

\markboth{IEEE Transactions on Visualization and Computer Graphics,~Vol.~xx, No.~x, Month~202x}{Chai \MakeLowercase{\textit{et al.}}: {\name}: Skeleton-aware Text-based 4D Avatar Generation with In-Network Motion Retargeting}

\maketitle

\begin{abstract}

The creation of 4D avatars ({\ie}, animated 3D avatars) from text description typically uses text-to-image (T2I) diffusion models to synthesize 3D avatars in the canonical space and subsequently applies animation with target motions. 
However, such an optimization-by-animation paradigm has several drawbacks. 
(1) For pose-agnostic optimization, the rendered images in canonical pose for na\"ive Score Distillation Sampling (SDS) exhibit domain gap and cannot preserve view-consistency using only T2I priors, and 
(2) For post hoc animation, simply applying the source motions to target 3D avatars yields translation artifacts and misalignment.
To address these issues, we propose \textbf{S}keleton-aware \textbf{T}ext-based 4D \textbf{A}vatar generation with in-network motion \textbf{R}etargeting ({\name}). 
{\name} considers the geometry and skeleton differences between the template mesh and target avatar, and corrects the mismatched source motion by resorting to the pretrained motion retargeting techniques. With the informatively retargeted and occlusion-aware skeleton, we embrace the skeleton-conditioned T2I and text-to-video (T2V) priors, and propose a hybrid SDS module to coherently provide multi-view and frame-consistent supervision signals. Hence, {\name} can progressively optimize the geometry, texture, and motion in an end-to-end manner.
The quantitative and qualitative experiments demonstrate our proposed {\name} can synthesize high-quality 4D avatars with vivid animations that align well with the text description.
Additional ablation studies shows the contributions of each component in {\name}. 
The source code and demos are available at: \href{https://star-avatar.github.io}{https://star-avatar.github.io}.

\end{abstract}


\section{Introduction}
\label{sec:intro}

\IEEEPARstart{R}{}ecent years have witnessed an impressive progress of text-to-image (T2I) generative models~\cite{rombach2022high,saharia2022photorealistic,karras2019style,nichol2021glide}. 
By leveraging the versatile T2I diffusion priors~\cite{rombach2022high,saharia2022photorealistic}, the creation of 3D content~\cite{poole2023dreamfusion,lin2023magic3d,zhang2024text2nerf,liu2023zero,szymanowicz2024splatter_image} (especially human-like characters~\cite{liao2024tada,kolotouros2023dreamhuman,huang2023humannorm,xu2023seeavatar,cao2023guide3d}) from arbitrary text description have received great attention and interest in the intersection of computer vision and graphics communities.

\begin{figure}[t!]
    \centering
          \begin{overpic}[trim=4cm 0cm 3cm 0cm,clip,width=1\linewidth,grid=false]{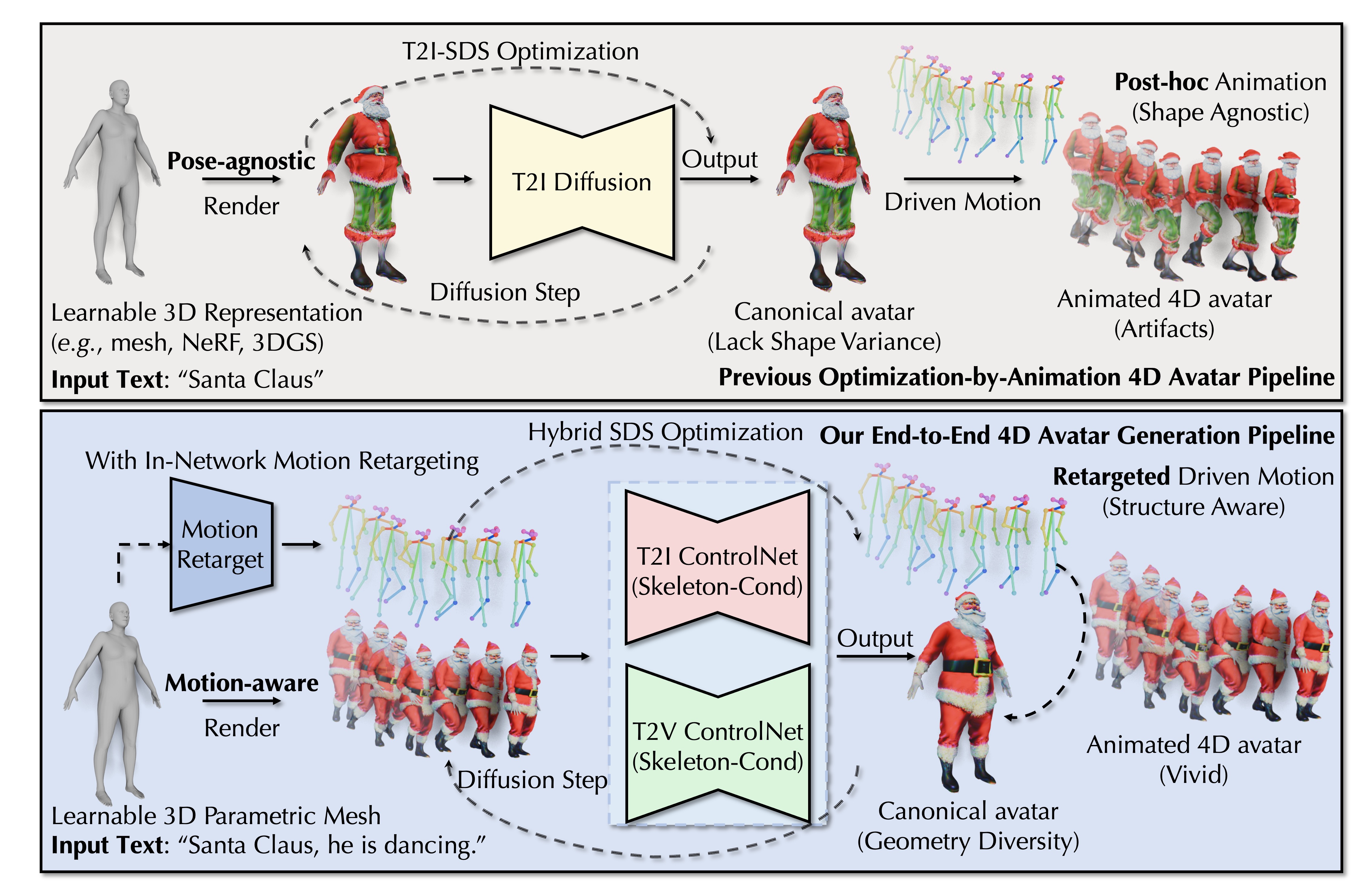}
      \end{overpic}
    \caption{\textbf{Typical Optimization-by-Animation Pipeline (\textit{Top}) {\vs} Ours (\textit{Bottom}) for Text-based 4D Avatar Generation.} \textit{Top}. Previous methods ({\eg}, TADA~\cite{liao2024tada}) first optimize the 3D representation in fixed canonical pose and subsequently apply motions for animation, which lack \textit{shape diversity} and exhibit \textit{animation artifacts}. \textit{Bottom}. The proposed {\name} for 4D avatar generation. It leverages in-network motion retargeting and hybrid SDS to jointly update geometry, texture, and motions to achieve vivid 4D avatar from only text description.}
    \label{fig.pipecmp}
\end{figure}

In an extended application, the creation of 4D avatar -- synthesizing animatable characters with authentic human motions -- has been found valuable in film and gaming industry. 
Existing text-based 4D avatar generation follows a typical optimization-by-animation paradigm (see Fig.~\ref{fig.pipecmp}).
The state-of-the-art methods~\cite{liao2024tada,huang2023dreamwaltz,liu2023humangaussian,Jiang_2023_ICCV} first resort to the use of image priors~\cite{rombach2022high} for generating 3D avatars from a text description, and subsequently apply the desired human motion from audio~\cite{tseng2023edge,li2023audio2gestures,yi2023generating}, video~\cite{cai2024smpler,lin2023one}, or text~\cite{jiang2023motiongpt,zhang2024motiondiffuse,gao2024guess} to generate 4D avatars.
However, we observe that such a typical paradigm encounters several challenges.

\begin{figure*}[t!]
    \centering
    \begin{overpic}[trim=15cm 0cm 10cm 0cm,clip,width=1.\linewidth,grid=false]{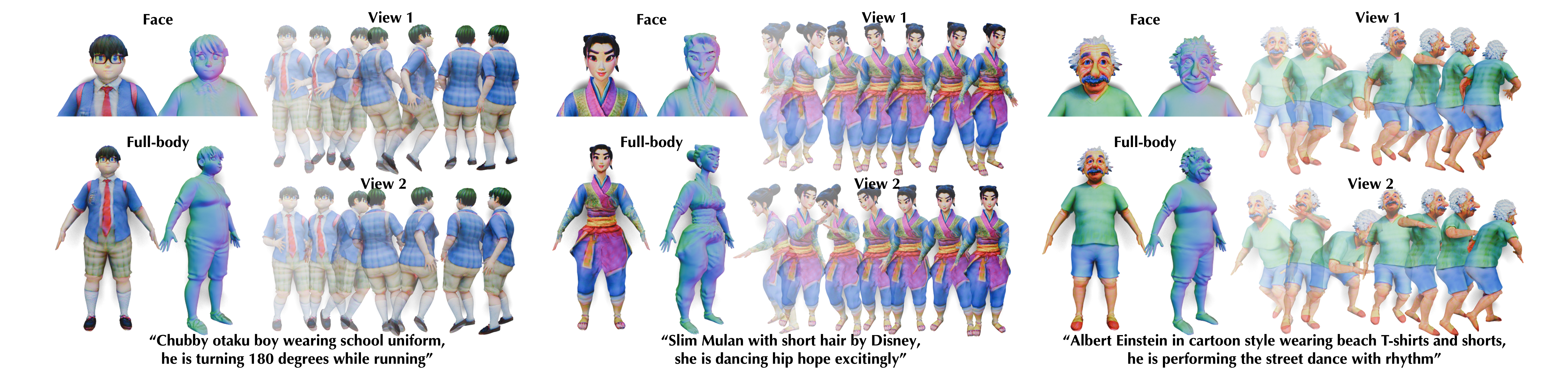} \end{overpic}
    \begin{overpic}[trim=15cm 0cm 10cm 0cm,clip,width=1.\linewidth,grid=false]{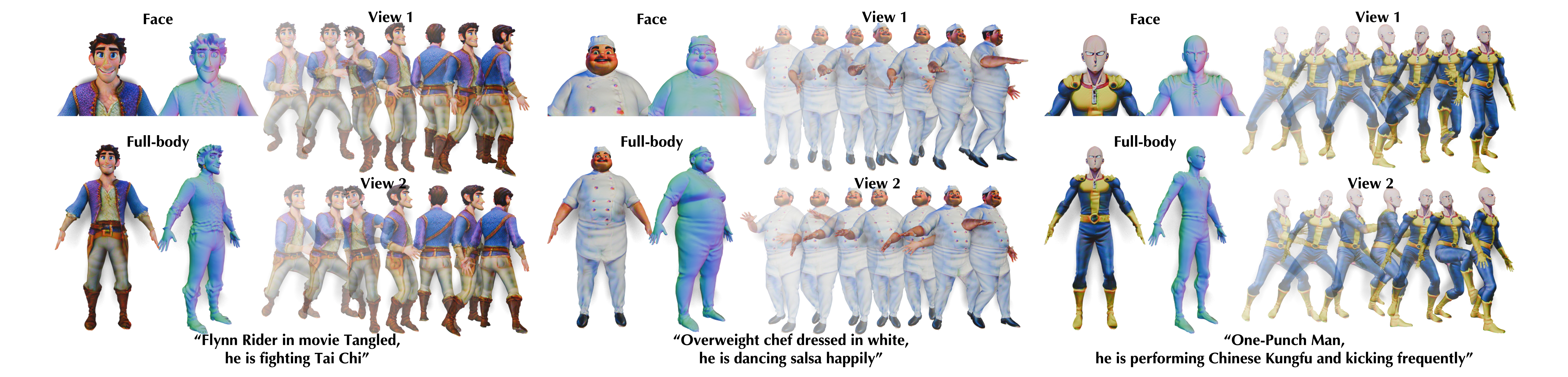} \end{overpic}
    \caption{
    \textbf{Examples of Generated 4D Avatars.} We propose {\name} to produce high-fidelity 4D avatars from only text description. For each sample, we showcase: \textit{Left}. Face and full-body of the textured 3D avatar and its normal map, and \textit{Right}. Randomly selected frames of the 4D animation in two different camera views. Best viewed in color and zoom-in.}
    \label{fig.teaser}
\end{figure*}

First, existing methods~\cite{liao2024tada,xu2023seeavatar,Jiang_2023_ICCV,zhang2023avatarverse} fail to address the pose distribution difference between the rendered 2D avatar in canonical space and that of T2I models. 
This is because T2I priors are trained in a majority of natural human poses. 
Therefore, such canonical pose-driven optimization suffers from a potential domain gap and is thus detrimental to obtaining expressive 3D avatars.
In contrast, while several methods~\cite{kolotouros2023dreamhuman,huang2023dreamwaltz,zhang2023avatarstudio} learn deformable 3D representations~\cite{alldieck2021imghum} with sampled pose~\cite{pavlakos2019expressive}, they implicitly suggest the source template and target avatar posses similar geometry and skeleton structures. 
Thus, it requires additional regularization to constrain the 3D representations to achieve reasonable deformations, which sacrifices the geometry diversity.
Second, the optimization step based on T2I Score Distillation Sampling (SDS) lacks stability and may potentially deteriorate into local optimum. 
The reason is that generated identity and/or appearance with different viewpoints exhibit frame inconsistency and content drift from na\"ive T2I priors, {\eg},~inconsistent background, appearance, and illumination.
Finally, due to the skeleton and geometry difference between the source template and target avatar, animation artifacts ({\eg}, body intersection) happen when simply copying the source motion from motion generative models~\cite{yi2023generating,jiang2023motiongpt,zhang2024motiondiffuse} to create a target 4D avatar.

To address the above challenges, we propose a \textbf{S}keleton-aware \textbf{T}ext-based 4D \textbf{A}vatar Generation with in-network motion \textbf{R}etargeting ({\name}) model.
{\name} seamlessly synthesizes high-fidelity 4D avatar with authentic animations from only text description (see Fig.~\ref{fig.teaser}). 
Specifically, the optimization process of {\name} is skeleton-aware, geometry-aware, and motion-aware. 
{\name} progressively optimizes the geometry, appearance, and motion such that it compounds realistic 4D avatar to align the text description, with three crucial insights beyond prior art. 
First, inspired by~\cite{zhang2024motiondiffuse,zhang2023skinned}, we perform in-network motion retargeting of the initialized motions. 
As a result, the updated motions and skeleton possess precise structural information of the target 3D avatar while maintaining the semantics of the given text description. 
Second, to eliminate the disturbance of temporal-inconsistency and view-agnostic SDS with na\"ive T2I models, we resort to the training-free text-to-video (T2V) diffusion models~\cite{peng2023conditionvideo} to provide frame-consistency priors. By coherently conditioning on the retargeted, occlusion-aware skeleton with ControlNet~\cite{zhang2023adding}, {\name} provides informative skeleton conditions that contributes to high-quality avatars of the optimization stage.
Finally, to improve the geometry-appearance alignment among different views and poses, we propose a hybrid SDS, which is a compound of multi-view SDS and masked sequential SDS, as well as accompanied with the elaborately designed hierarchical regularization, to train the 3D representation that stabilizes the optimization process and achieve satisfactory 4D avatar.

The aforementioned insights and techniques successfully address the challenges of text-based 4D avatar creation. 
The proposed {\name} seamlessly generates vivid 4D avatars from only text description, and is flexible to deploy with arbitrary motions generated from either text, audio, or videos, while preserving the motion transfer quality of the 4D avatar.
Experimental results demonstrate {\name} synthesizes high-fidelity 4D avatar which aligns well with the given text description. Both quantitative and qualitative experiments demonstrate our {\name} achieves state-of-the-art generation results compared to previous methods with the optimization-by-animation pipeline.

In summary, our main contributions are as follows:
\begin{itemize}
    \item To the best of our knowledge, we are the earliest work to analyze the influence of missing and/or mismatched motion in text-to-avatar generation. We propose a novel end-to-end model {\name} to address these drawbacks in 4D avatar generation.
    \item We systemically design the {\name} pipeline with integration of text-to-motion, motion retargeting, and text-to-3D techniques with crucial modifications. This enables {\name} to generate high-quality avatars with vivid animations from text descriptions.
    \item We combine the capabilities of skeleton-conditioned T2I and T2V priors, equipped with occlusion-aware, coarse-to-fine hybrid SDS, and hierarchical regularization to improve the optimization process and address the view-inconsistency and facial-blur issues.
    \item Quantitative and qualitative experiments validate {\name} can generate vivid 4D avatars and outperform the optimization-by-animation baselines. Code and trained models will be released for research purposes. 
\end{itemize}

\section{Related Work}

The tasks of 3D content generation with text description have received increasing attention in recent years. 
A feasible solution is to leverage the 2D models prior~\cite{rombach2022high,saharia2022photorealistic,nichol2021glide,ramesh2022hierarchical} pretrained on $\langle \textit{image}, \textit{text} \rangle$ pairs to optimize the 3D generation.
This section reviews methods that are related to this work. 
Comprehensive surveys can be found in~\cite{li2023generative,lee2024text,liu2024comprehensive}.

\myparagraph{Text-driven 3D Content Generation.} 
The seminal works ({\eg}, DreamField~\cite{jain2022zero}, CLIP-mesh~\cite{mohammad2022clip}, AvatarCLIP~\cite{hong2022avatarclip}, CLIP-Face~\cite{aneja2023clipface}, Text2Mesh~\cite{michel2022text2mesh}) optimize the underlying 3D representation ({\eg}, NeRF~\cite{mildenhall2020nerf}, 3DGS~\cite{kerbl20233d}, or mesh) via CLIP~\cite{radford2021learning} space similarity between rendered images and text embeddings. 
However, these methods exhibit less realism since CLIP only provides high-level semantics.
Recent advances resort to the versatile T2I diffusion~\cite{rombach2022high} priors for text-guided 3D generation. The pioneering work, DreamFusion~\cite{poole2023dreamfusion} and SJC~\cite{wang2023score}, propose Score Distillation Sampling (SDS) for text-to-3D generation. Follow-up works such as ProlificDreamer~\cite{wang2024prolificdreamer}, PGC~\cite{pan2024enhancing}, DreamTime~\cite{huang2024dreamtime} revisit and improve the na\"ive SDS for better generation quality. One line of work improves the training pipeline with elaborately designed branches. Magic3D~\cite{lin2023magic3d} proposes a coarse-to-fine pipeline to generate fine-grained 3D meshes. Fantasia3D~\cite{chen2023fantasia3d} disentangles the generation of geometry and texture, Guide3D~\cite{cao2023guide3d}, MVDream~\cite{shi2024mvdream}, MVD-Fusion~\cite{hu2024mvdfusion} and TEXTure~\cite{richardson2023texture} leverage the multi-view images or features for improving view-consistency.
More recently, by leveraging the hybrid text-driven image~\cite{rombach2022high,saharia2022photorealistic,podell2024sdxl}, video~\cite{blattmann2023align,guo2024animatediff,ho2022imagen,zhou2022magicvideo}, and 3D-aware~\cite{liu2023zero,shi2024mvdream,liu2024syncdreamer} diffusion models, MAV3D~\cite{singer2023text}, 4D-fy~\cite{bahmani20234d}, and AYG~\cite{ling2023align} focus on generate 4D dynamic scenes using deformable 4D representations~\cite{muller2022instant,cao2023hexplane,luiten2023dynamic}.

\myparagraph{Text-driven 3D Avatar Generation.}
By integrating explicit~\cite{pavlakos2019expressive,loper2015smpl} and/or implicit~\cite{alldieck2021imghum,muller2022instant,wang2021neus} 3D human priors for full-body 3D human creation from text description, early works such as AvatarCLIP~\cite{hong2022avatarclip} and CLIP-Actor~\cite{youwang2022clipactor} optimize the 3D representations~\cite{mildenhall2020nerf,wang2021neus} through CLIP similarly~\cite{radford2021learning}. Due to the limited guidance from CLIP, these models exhibit noisy geometry and blurry texture.
Recent advances~\cite{liao2024tada,kolotouros2023dreamhuman,xu2023seeavatar,Jiang_2023_ICCV,huang2023avatarfusion,yuan2023gavatar} integrate the power of diffusion models with SDS and achieve promising progress with reasonable details of text guided 3D avatar generation, which follow the similar paradigm of text-to-3D content generation~\cite{poole2023dreamfusion}.
AvatarCraft~\cite{Jiang_2023_ICCV} and DreamHuman~\cite{kolotouros2023dreamhuman} leverage the pretrained T2I models and SMPL~\cite{loper2015smpl} body prior to train implicit 3D representations~\cite{alldieck2021imghum,mildenhall2020nerf,wang2021neus}, with coarse-to-fine and semantic zoom techniques. TADA~\cite{liao2024tada} uses SMPL-X~\cite{pavlakos2019expressive} model and learns explicit mesh for avatar generation.
To improve the optimization process with more complex avatars, state-of-the-art methods resort to ControlNet~\cite{Zhang_2023_ICCV} with elaborately designed branches to provide view-consistent guidance such that improves the plausible details. Existing methods uses skeleton ({\eg}, DreamWaltz~\cite{huang2023dreamwaltz} and HumanGaussian~\cite{liu2023humangaussian}), keypoint ({\eg}, DreamAvatar~\cite{cao2023dreamavatar} and AvatarBooth~\cite{zeng2023avatarbooth}), DensePose~\cite{guler2018densepose} ({\eg}, AvatarVerse~\cite{zhang2023avatarverse} and AvatarStudio~\cite{zhang2023avatarstudio}) or normal/depth ({\eg}, HumanNorm~\cite{huang2023humannorm})
as conditions with modified SDS.
A common trend of the above work is either to use a fixed canonical pose or a sampled template pose for optimization. As a result, it is challenging to generate personalized structures and high-fidelity avatars.

\myparagraph{Character Animation \& Motion Retargeting.}
One line work of text-driven avatar generation focuses on generating animatable 3D avatars~\cite{liao2024tada,huang2023dreamwaltz,Jiang_2023_ICCV,youwang2022clipactor} for 4D content.
These methods typically apply the prepared motions~\cite{yi2023generating,jiang2023motiongpt,petrovich2021action,tevet2023human,yuan2020dlow,shafir2023human} to animate the canonical avatar for 4D content creation. For example, by leveraging the text-driven human generation models~\cite{petrovich2022temos,guo2022generating,ahuja2019language2pose}, which synthesize realistic human-like motion from textual descriptions, TADA~\cite{liao2024tada} can directly animate the avatars for its explicit mesh representation, while AvatarCraft~\cite{Jiang_2023_ICCV}, DreamWaltz~\cite{huang2023dreamwaltz} and HumanGaussian~\cite{liu2023humangaussian} use additional deformable modules to translate or query the motion deformation into corresponding implicit representations.
However, these methods neglect the shape and skeleton difference between the template shape and target avatar, which thereby introduce artifacts if simply copy the source motion for avatar animation. In the field of computer graphics, motion retargeting~\cite{retargettingretargetting,kim2016retargeting,pan2024expressive,ji2023stylevr} is a long-standing space-time optimization problem that identifies features of the source motions as kinematic constraints. 
To improve the retargeting quality, recent works use deep learning techniques~\cite{zhang2023skinned,lim2019pmnet,villegas2018neural,aberman2020skeleton,hu2023pose} to learn motion representations from skeleton and/or geometry constraints.

To the best of our knowledge, while there have been impressive progress in the area of text-based generation, motion retargeting has not been investigated to address the animation artifacts for text-to-avatar generation. In the next section, we embrace skeleton-aware and geometry-aware motion retargeting to produce 4D avatars.

\section{Method}

\subsection{Preliminary}

\myparagraph{Avatar Representation.} 
In Eq.~\ref{eq.smplx}, we use the re-topologized SMPL-X~\cite{pavlakos2019expressive} with per-vertex displacement $\delta$ and UV texture $\alpha$ to represent a textured 3D avatar $\boldsymbol{m}(\beta,\psi,\delta;\boldsymbol{q},\alpha)$.
    \begin{align}
        \boldsymbol{m}(\beta,\psi,\delta; \boldsymbol{q}, \alpha) & = \mathcal{W}\left( \bf{T}(\beta,\psi,\delta; \boldsymbol{q}), \mathcal{J}(\beta); \boldsymbol{q}, \omega,\alpha \right) \label{eq.smplx} \\
        \bf{T}(\beta,\psi,\delta;\boldsymbol{q}) &= \bar{\boldsymbol{m}} + \boldsymbol{b}_s(\beta) + \boldsymbol{b}_e(\psi) + \boldsymbol{b}_p(\boldsymbol{q})+ \delta,
    \end{align}
where $\beta$ and $\psi$ represent the shape and expression parameters. $\boldsymbol{q}$ indicates the sampled pose from the motions $\mathcal{Q}$. 
$\mathcal{W}$ represents linear blending skinning function~\cite{lewis2023pose}, with predefined blend weights $\omega$. $\mathcal{J}$ represents the 3D joint location regressor. 
$\bar{\boldsymbol{m}}$ is the shape template, $\boldsymbol{b}_s, \boldsymbol{b}_e$ and $\boldsymbol{b}_p$ are shape, expression, and pose bases, respectively.
In this work, we denote the learnable parameters as $\Theta \coloneqq \{\beta, \psi, \delta, \alpha\}$ for simplicity.

\myparagraph{Score Distillation Sampling}
(SDS)~\cite{poole2023dreamfusion} leverages the pretrained 2D diffusion model to minimize the difference between the predicted noise $\epsilon_{\phi}(\boldsymbol{x}_t;y,\tau)$ and Gaussian noise $\epsilon \sim \mathcal{N}(\mathbf{0},\mathbf{I})$, such that it computes gradient to optimize the learnable parameter $\Theta$:
\begin{equation}
    \nabla_{\Theta} \mathcal{L}_{\text{SDS}}(\phi,\boldsymbol{x}) = \mathbb{E}_{\tau,\epsilon}\left[w(\tau)(\epsilon_{\phi}(\boldsymbol{x}_\tau;y,\tau) - \epsilon )\frac{\partial \boldsymbol{x}}{\partial \Theta } \right],
    \label{eq.SDS}
\end{equation}
where $\boldsymbol{x} \coloneqq g(\Theta; \boldsymbol{q}, \boldsymbol{\pi})$ represents the differentiable rendering of the 3D model, given a body pose $\boldsymbol{q}$ and camera view $\boldsymbol{\pi}$. $y$ indicates the text condition, $w(\tau)$ is the weight function correlated to noise level $\tau$.

\begin{figure*}
    \centering
         \begin{overpic}[trim=5cm 6.5cm 1cm 0cm,clip,width=1\linewidth,grid=false]{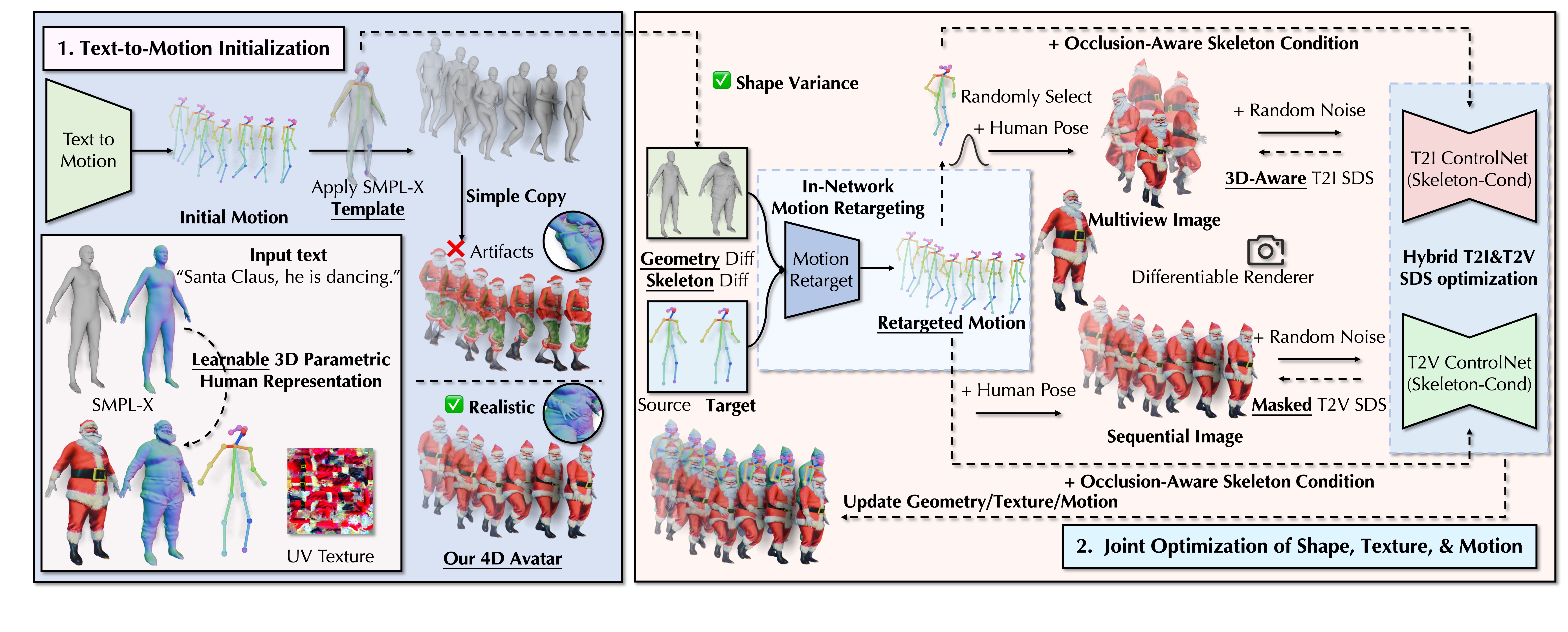} \end{overpic}
    \caption{\textbf{Overview of the Proposed {\name}.} \textit{Left}. Given a text description, we initialize the human motion with pretrained text-to-motion model~\cite{zhang2024motiondiffuse}. Note that the typical optimization-by-animation paradigm easily yields deteriorated body structures and animation artifacts for 4D avatar generation. \textit{Right}. We eliminate the potential pose distribution bias in the SDS-based optimization by integrating the retargeted motion for animation. With the personalized and occlusion-aware skeleton, we leverage the hybrid T2I and T2V diffusion models to provide 3D consistent priors that progressively optimize the geometry, texture, and motion to produce 4D avatar in an end-to-end manner.}
    \label{fig.pipeline}
\end{figure*}

\subsection{Motivation and Overview}
Previous methods typically exhibit an optimization-by-animation paradigm to create 4D avatars, which firstly resort to SDS~\cite{poole2023dreamfusion} in Eq.~\ref{eq.SDS} or its variants~\cite{huang2023dreamwaltz,zhang2023avatarverse,zhang2023avatarstudio} to train the 3D representation by performing the denoising process of rendered canonical avatars. 
However, the diffusion models are pretrained to estimate images with diverse poses, which exhibit pose-distribution bias in the optimization stage and hinder the generation fidelity.
In the animation stage, we notice that simply assigning the motions to the canonical avatar yields artifacts, as it ignores the skeleton and geometry difference between the source template and the target avatar. These two key issues make the text-to-avatar models unable to generate vivid 4D avatars with both shape diversity and motion quality.

To mitigate these issues, we propose the \textbf{S}keleton-aware \textbf{T}ext-based 4D \textbf{A}vatar generation with in-network motion \textbf{R}etargeting model ({\name}). Fig.~\ref{fig.pipeline} shows the overview pipeline of the proposed method, which contains three crucial components. 
\textbf{First}, the optimization process is skeleton-aware and motion-aware. By leveraging the sampled human motions generated from text descriptions~\cite{zhang2024motiondiffuse}, we make the rendered 3D avatar align better with natural pose distribution in diffusion priors. 
\textbf{Second}, to address the artifacts when simply copying the source motions into target geometry, we employ in-network motion retargeting~\cite{zhang2023skinned} to progressively update the initial motions such that the motions faithfully reflect the personalized structure of the target avatar during the training stage. 
\textbf{Third}, to improve the optimization consistency, instead of na\"ive T2I-based SDS, we propose a hybrid SDS approach that employs both T2I~\cite{rombach2022high} and T2V~\cite{peng2023conditionvideo} priors with occlusion-aware, skeleton-conditioned ControlNet~\cite{zhang2023adding} to provide strong view-consistency supervisions and improve the generation quality. 
\mbox{The following sections describe each component in detail.}

\subsection{Skeleton-Aware Text-based 4D Avatar Generation}

\myparagraph{Motion Initialization.}
To synthesize an initial motion that aligns with the given text description, we leverage a pretrained text-to-motion model $\mathcal{M}$~\cite{zhang2024motiondiffuse} to translate the text description into motion sequences $\mathcal{Q}^{s} = \{\boldsymbol{q}^{s}_0,\cdots, \boldsymbol{q}^{s}_\ell \}$ with length $\ell =196$.
However, the initial motion is based on a source template mesh $\boldsymbol{m}^s$ with joints $\boldsymbol{j}^s$. It exhibits geometry and skeleton mismatch when we simply copy $\mathcal{Q}^s$ to animate an avatar $\boldsymbol{m}^t$.

\myparagraph{In-Network Motion Optimization.}
Intuitively, sampling and rendering avatars with random poses for training can potentially eliminate the distribution bias such that it improves the synthesized quality and optimization stability. 
However, we observe that simply applying a pose $\boldsymbol{q}^s_i$ to the target avatar yields the intersection artifacts of body regions, especially for those with shape-variant bodies. As a result, it leads to counterproductive drawbacks for motion-aware optimization (as demonstrated in Sec.~\ref{sec.abl}).

To simultaneously eliminate motion artifacts and pose inconsistency, and improve the generation quality, we leverage a pretrained motion retargeting model {\small $\mathcal{F} = \{\Delta{\mathcal{F}}_{\boldsymbol{m}}, \Delta{\mathcal{F}}_{\boldsymbol{j}}\}$}~\cite{zhang2023skinned} to correct the initial motion conditioned by both skeleton and geometry of a target avatar $\boldsymbol{m}^t$. At optimization step $T$, we retarget the source motion $\mathcal{Q}^{s}$ to the target motion $\mathcal{Q}^t_T$ conditioned by the canonical mesh $\boldsymbol{m}^{*t}_T$ and its joints $\boldsymbol{j}^{*t}_T$, with geometry-aware and skeleton-aware residuals $\delta \mathcal{Q}_T$, 
\begin{equation}
   \mathcal{Q}^t_T = \mathcal{Q}^{s} + \underbrace{\Delta{\mathcal{F}}_{\boldsymbol{m}}(\boldsymbol{m}^s, \boldsymbol{m}^{*t}_T) + \Delta{\mathcal{F}}_{\boldsymbol{j}}(\boldsymbol{j}^s, \boldsymbol{j}^{*t}_T)}_{\delta \mathcal{Q}_T}.
    \label{eq.motion_retarget}
\end{equation}

In this way, we obtain the animated geometry $\boldsymbol{m}^t_T$ and joints $\boldsymbol{j}^t_T$ using the retargeted motion $\mathcal{Q}^t_T$. The 3D geometry and motions {\wrt} the text description are updated progressively during optimization, which in return, provide precisely rendered images with diverse motions for the diffusion process.

\subsection{Hybrid SDS with Frame Consistency}

\myparagraph{Skeleton-conditioned ControlNet with Occlusion Detection.}
Vanilla SDS~\cite{poole2023dreamfusion} leverages view-dependent prompts for text-to-image diffusion models to provide sparse supervision signals. Thus, the generation results tend to be imprecise and will potentially introduce Janus (multi-face) problems (see Fig.~\ref{fig.canocmp}).
Inspired by the recent success of ControlNet~\cite{zhang2023adding} in text-to-avatar generation~\cite{huang2023dreamwaltz,liu2023humangaussian,zhang2023avatarverse,zhang2023avatarstudio}, we use the skeleton from retargeted human motions as precise conditions.

To eliminate the visibility ambiguity of the rendered skeleton during the diffusion process, we leverage the informative rasterization process in differentiable rendering~\cite{johnson2020accelerating} to select visible skeleton $\boldsymbol{j}^{o}$. Specifically, given camera view $\boldsymbol{\pi}$, we denote the vertices of $\boldsymbol{m}^t$ as $\boldsymbol{v}^t$ and the $i$-th joints $\boldsymbol{j}^t_i$ are visible {\iff} $\boldsymbol{j}^t_i$'s $k$-nearest neighbor is located in the visible vertices $\boldsymbol{v}' \subseteq \boldsymbol{v}$, indicated by rasterization step in rendering.
Then, we leverage the occlusion-aware skeleton $\boldsymbol{j}^o$ as condition for ControlNet~\cite{zhang2023adding} to compute the revised T2I SDS as
\begin{equation}
    \begin{aligned}
        \nabla_{\Theta}& \mathcal{L}_{\text{SDS}}(\phi,\boldsymbol{x}^*)  = \\ & \mathbb{E}_{\tau,\epsilon}\left[\omega(\tau)\left(\epsilon_{\phi}(\boldsymbol{x}^*_\tau;\boldsymbol{j}^o,y,\tau) - \epsilon \right)\frac{\partial \boldsymbol{x}^*}{\partial \Theta } \right],
    \end{aligned}
    \label{eq.skelsds}
\end{equation}
where $\boldsymbol{x}^* \coloneqq g(\Theta; \boldsymbol{q}^*, \boldsymbol{\pi})$ indicates the rendered images with retargeted motions $\boldsymbol{q}^*\in \mathcal{Q}^t_T$ at training step $T$. 

\myparagraph{Masked Sequential SDS.}
While the skeleton-conditioned ControlNet can provide precise view-dependent knowledge, we observe that diffusion models potentially exhibit instability across different views.
To improve the frame-consistency,
apart from leveraging the pretrained T2I models and the given retargeted motion sequence $\mathcal{Q}^t$, we additionally resort to the training-free T2V models~\cite{peng2023conditionvideo} to present frame-consistent priors of constitute views during the 4D avatar creation steps. 
However, empirical observation shows that while the T2V model~\cite{peng2023conditionvideo} can achieve plausible consistency in the body parts, it yields blurry regions in the fine-grained structures ({\eg}, the face regions). Therefore, we select facial mask and propose a masked sequential SDS by eliminating the potential error on the face regions. Specifically, we obtain the non-facial mask $\boldsymbol{x}^*_{\text{mask}}$ in rendering of given camera view $\boldsymbol{\pi}$ and eliminate the facial regions for sequential SDS, as Eq.~\ref{eq.seqsds} shows.
\begin{equation}
    \begin{aligned}
        \nabla_{\Theta} & \mathcal{L}^{\text{seq}}_{\text{SDS}}(\phi,\boldsymbol{x}^*, \boldsymbol{x}^*_{\text{mask}}; \mathcal{Q}' \subseteq \mathcal{Q}^t) = \\
        & \mathbb{E}_{\mathcal{Q}',\tau,\epsilon}\left[ \boldsymbol{x}^*_{\text{mask}} \odot \omega(\tau)\left(\epsilon^{*}_{\phi}(\boldsymbol{x}^*_\tau;\boldsymbol{j}^o,y,\tau,l) - \epsilon \right)\frac{\partial \boldsymbol{x}^*}{\partial \Theta } \right], 
    \end{aligned}
    \label{eq.seqsds}
\end{equation}
where $\mathcal{Q}'$ is the sampled $l$-length motion from $\mathcal{Q}^t$ for T2V model, with noise predictor $\epsilon^{*}_{\phi}(\cdot)$ modified from T2I model $\epsilon_{\phi}(\cdot)$, $\odot$ indicates Hadamard product.

\subsection{Hierarchical Regularization}
To prevent the model from learning noisy geometry, we propose simple yet effective regularization terms, namely hierarchical geometry regularization, to improve both the global structure and the fine-grained local details of 3D geometry. Specifically, $\mathcal{L}_{\text{shp}}$ in Eq.~\ref{eq.betareg} regulates the shape parameters $\beta$ to ensure a reasonable global structure of the generated avatar.
\begin{equation}
    \mathcal{L}_{\text{shp}} = \left \| \beta  \right \| _2^2.
    \label{eq.betareg}
\end{equation}

In addition, to improve the smoothness of geometry details from per-vertex displacement $\delta$, we use Laplacian smoothness $\mathscr{L}(\boldsymbol{m}^t)$ to regulate $\delta$ as Eq.~\ref{eq.reglap} illustrates.
\begin{equation}
   \mathcal{L}_{\text{lap}} = \left \| \boldsymbol{m}^t - \mathscr{L}(\boldsymbol{m}^t)  \right \| _2^2.
    \label{eq.reglap}
\end{equation}

In addition to the aforementioned global constraints, we observe that SDS can be sensitive to facial structures, especially the eye and mouth regions. Therefore, in Eq.~\ref{eq.reglmk}, we use the predefined facial keypoints to preserve the relative positions and penalize the intersections of eye and lip contours, such that regulate $\boldsymbol{m}^t$ with reasonable facial structure.
\begin{equation}
    \begin{aligned}
    \mathcal{L}_{\text{face}} &= \mathds{1}\left(\delta (v^*_u, v^*_l)<0\right)||v^*_u- v^*_l||^2_2\\ 
    & + \mathds{1}(||v^*_e - v^*_f||^2_2 < r^2 )||v^*_e - v^*_f||^2_2,      
    \end{aligned}
    \label{eq.reglmk}
\end{equation}
where $v^*_u, v^*_l$ indicate the paired upper-and-lower lip vertices, $v^*_e, v^*_f$ indicate vertices of eyeball and forehead regions, respectively, $r$ indicates the eyeball radius. $\mathds{1}(\cdot)$ is the indicate function that penalizes {\iff} $v^*_u$ lies below the lower lip $v^*_l$ or $v^*_e$ intersected with $v^*_f$.

Finally, the overall hierarchical geometry regularization $\mathcal{L}_{\text{reg}}$ is weighted summarize of the above terms, 
\begin{equation}
    \mathcal{L}_{\text{reg}} = \mathcal{L}_{\text{shp}} + \mathcal{L}_{\text{lap}} + \mathcal{L}_{\text{face}},
    \label{eq.regfinal}
\end{equation}
where we eliminate the weights of each term for clarity. The weights are empirically set to balance the smoothness and diversity, and keep consistent for different text prompts in our experiments.

\subsection{Overall Optimization Algorithm}

\begin{algorithm}[t!]
\caption{Training pipeline of {\name} for text-to-4D avatar generation.}
\label{alg.train}
\SetKwInOut{Input}{Input}
\SetKwInOut{Return}{Return}
\SetKwInOut{Require}{Require}
\SetKwInOut{Output}{Output}
\SetKwComment{Comment}{\# }{}
\Input{Text description $y$}
\Require{
Learning rate $\eta$;
Pretrained T2I prior $\epsilon_\phi$;
Pretrained motion generation model $\mathcal{M}$;
Pretrained motion retargeting model $\mathcal{F}$;
Training step $T^*$
}
\Output{4D Avatar $\mathrm{M}^* \coloneqq \{\boldsymbol{m}| \boldsymbol{m}(\Theta^*_T; \boldsymbol{q}), \forall \boldsymbol{q}\in \mathcal{Q}^t_T\}$}
$\mathcal{Q}^t_0 = \mathcal{Q}_s \gets \mathcal{M}(y)$ \Comment*[r]{Motion Initialization}
\While{$T \leq T^*$}{

    Sample motion $\mathcal{Q}'_T \subseteq \mathcal{Q}^t_T$ and random pose $\boldsymbol{q}_T \in \mathcal{Q}^t_T$;

    Generate animated 4D avatar: \\ $\mathrm{M}_T \coloneqq \{\boldsymbol{m}| \boldsymbol{m}(\Theta_T; \boldsymbol{q}), \forall \boldsymbol{q}\in \mathcal{Q}'_T\}$ \Comment*[r]{Eq.~\ref{eq.smplx}} 

    Sample random camera view $\boldsymbol{\pi}$;
    
    Render video $\mathrm{X}_T = \{\boldsymbol{x} | \boldsymbol{x} \coloneqq g(\Theta_T; \boldsymbol{q}, c), \forall \boldsymbol{q}\in \mathcal{Q}'_T)\}$;
    
    Calculate gradient for T2V step, s.t. $\mathrm{M}_T, \mathrm{X}_T$: \\
    $\nabla_\Theta \mathcal{L}_{\text{T2V}} = \nabla_\Theta \mathcal{L}_{\text{SDS}}^{\text{seq}} + \nabla_\Theta \mathcal{L}_{\text{reg}}$ \Comment*[r]{Eq.~\ref{eq.seqsds} \& Eq.~\ref{eq.regfinal}}
    
    Update parameters $\Theta_{T} \gets \Theta_{T} - \eta \nabla_\Theta \mathcal{L}_{\text{T2V}}$;

    Sample $n$ camera views ${\Pi} = \{\boldsymbol{\pi}_1, \cdots, \boldsymbol{\pi}_n\}$;
    
    \While{$\boldsymbol{\pi} \in \Pi$}{ 
        Sample random pose $\boldsymbol{q}_T \in \mathcal{Q}^t_T$;
        
        Generate animated 3D avatar: \\
        $\boldsymbol{m}_T = \boldsymbol{m}(\Theta_T; \boldsymbol{q}_T)$  \Comment*[r]{Eq.~\ref{eq.smplx}}
        
        Render image $\boldsymbol{x}_T \coloneqq g(\Theta_T; \boldsymbol{q}_T , c)$;
        
        Calculate gradient for T2I step, s.t. $\boldsymbol{m}_T, \boldsymbol{x}_T$: \\
        $\nabla_\Theta \mathcal{L}_{\text{T2I}} = \nabla_\Theta \mathcal{L}_{\text{SDS}} + \nabla_\Theta \mathcal{L}_{\text{reg}}$ \Comment*[r]{Eq.~\ref{eq.skelsds} \& Eq.~\ref{eq.regfinal}}
        
        Update parameters $\Theta_{T} \gets \Theta_{T} - \eta \nabla_\Theta \mathcal{L}_{\text{T2I}}$;

    }
    Update motion $\mathcal{Q}^t_T \gets \mathcal{Q}^s + \delta \mathcal{Q}_T$, s.t. $\Theta_T$  \Comment*[r]{Eq.~\ref{eq.motion_retarget}}
    $T \gets T+1$;
}
\Return{$\mathrm{M}^* \coloneqq \{\boldsymbol{m}| \boldsymbol{m}(\Theta^*_T; \boldsymbol{q}), \forall \boldsymbol{q}\in \mathcal{Q}^t_T\}$.}

\end{algorithm}

The overall training pipeline is illustrated in Algo.~\ref{alg.train}. The proposed {\name} embraces the capability of T2I and T2V priors, with crucial motion-driven, occlusion-aware, and structure-aware information to provide multi-view and frame-consistency supervision signals for the SDS process. In this way, {\name} generates high-quality 4D avatar with vivid animations that faithfully aligns with text descriptions.

\section{Experiments}
\label{sec:experiment}

\subsection{Implementation Details}

We implement {\name} in PyTorch~\cite{paszke2019pytorch} and train the model on a single NVIDIA A100 GPU with 40GB memory. For the SDS guidance, we use Stable Diffusion v1.5~\cite{rombach2022high} as T2I and T2V priors, with classifier free guidance~\cite{ho2021classifierfree} set to $100$. 
We follow~\cite{liao2024tada} to sample camera poses for both head and body. In each training step, we leverage the full-body with random camera views for the T2V process to guarantee the global structure. We also randomly sample heads or bodies for the T2I process to improve the fidelity of facial appearances.
We use Adam optimizer~\cite{kingma2014adam} to train the model for $15,000$ steps. We set $k$-nearest neighbor as $k=20$ and $50$ for face/body occlusion detection.

\subsection{Quantitative Comparison}

\myparagraph{Experiment Setting.} We compare {\name} with the state-of-the-art (SOTA) methods~\cite{liao2024tada,huang2023dreamwaltz,liu2023humangaussian,Jiang_2023_ICCV}.
Specifically, we use 26 representative prompts varying from cartoon characters and realistic humans for quantitative comparisons.
For each prompt, we obtain the ``A-pose'' avatars of all evaluated methods.
For the 4D avatars, {\name} seamlessly generates the 4D avatar end-to-end from the text description.
For previous optimization-by-animation methods, we use the motion generated from pretrained text-to-motion models~\cite{zhang2024motiondiffuse} to produce animated results.

\myparagraph{Evaluation Metrics.}
We quantitatively evaluate all methods using the CLIPScore~\cite{hessel2021clipscore} (based on CLIP ViT-B/16) and VQAScore~\cite{lin2024evaluating} (based on CLIP-FlanT5-XXL).
CLIPScore evaluates the cosine similarity between the rendered images and the text descriptions, while VQAScore defines the alignment score with VQA probability ({\ie}, ``\texttt{Yes}'' answer to ``\texttt{Does this figure show \{text\}?}'' question).
Specifically, we evaluate the \textbf{avatar fidelity} by comparing scores between the \textit{object text} and rendered the canonical avatar from $8$ evenly distributed horizontal views, and the \textbf{animation fidelity} by comparing per-frame scores between videos and the text description (including motion description).

\begin{table}[]
\caption{\textbf{Quantitative Comparisons of the Canonical 3D Avatar and 4D Avatar.} We use CLIPScore and VQAScore metrics, higher score indicates better text-3D alignment. }
\setlength{\tabcolsep}{2pt}
\resizebox{1\linewidth}{!}{%
\begin{tabular}{l|cc|cc}
\toprule
\multicolumn{1}{l|}{\multirow{3}{*}{Evaluation Metric}} & \multicolumn{2}{c|}{Canonical 3D Avatar} & \multicolumn{2}{c}{4D Avatar}  \\
\cmidrule(lr){2-3} \cmidrule(lr){4-5} 
 & \multicolumn{1}{c}{CLIPScore$\uparrow$} & \multicolumn{1}{c|}{VQAScore$\uparrow$} & \multicolumn{1}{c}{CLIPScore$\uparrow$} & \multicolumn{1}{c}{VQAScore$\uparrow$} \\ \midrule
AvatarCraft~\cite{Jiang_2023_ICCV} & 32.09$\pm$2.83 & 0.56$\pm$0.15 & 32.66$\pm$2.16	 & 0.49$\pm$0.14 \\
DreamWaltz~\cite{huang2023dreamwaltz} & 35.74$\pm$2.80 & 0.76$\pm$0.12 & 35.05$\pm$2.89	& 0.57$\pm$0.18 \\
TADA~\cite{liao2024tada} & 36.88$\pm$2.07 & 0.79$\pm$0.16 & 35.57$\pm$1.77	& 0.61$\pm$0.15  \\
HumanGaussian~\cite{liu2023humangaussian} & 34.97$\pm$1.93 & 0.79$\pm$0.06 & 34.87$\pm$2.02&	0.59$\pm$0.15   \\ \midrule
\textbf{{\name} (Ours)} & \textbf{37.28$\pm$2.09} & \textbf{0.83$\pm$0.10} & \textbf{36.42$\pm$2.52} & \textbf{0.62$\pm$0.13} \\ \bottomrule
\end{tabular}
}
    \label{tab.clipscore}
\end{table}

\begin{table}[]
    \centering
    \caption{\textbf{User Study Results.} We show the average percentage ($\%$) of each baseline and our {\name} being selected as ``best'' according to three aspects: Q1. Geometry Quality, Q2. Appearance Quality, and Q3. Motion Quality. }
    \setlength{\tabcolsep}{4pt}
\resizebox{1\linewidth}{!}{%
    \begin{tabular}{l|c|c|c}
    \toprule
    Methods & Geo. Qual.$\uparrow$  & App. Qual.$\uparrow$ & Mot. Qual.$\uparrow$ \\ \midrule
    AvatarCraft~\cite{Jiang_2023_ICCV} & 3.95  & 4.50 & 7.56 \\
    DreamWaltz~\cite{huang2023dreamwaltz} & 23.89  & 22.48 & 23.75 \\
    TADA~\cite{liao2024tada} & 17.36  & 17.20 & 16.77 \\
    HumanGaussian~\cite{liu2023humangaussian} & 11.32  & 12.65 & 12.43 \\ \midrule
    \textbf{{\name} (Ours)} & \textbf{43.48} & \textbf{43.17} &  \textbf{39.49} \\ \bottomrule
    \end{tabular}}
    \label{tab.user_study}
\end{table}

\myparagraph{User Study.}
Since CLIPScore and VQAScore only consider per-frame text-image similarity, these metrics are not able to estimate the visual quality in the temporal dimension. To further demonstrate the quality of {\name} for 4D avatar generation, we use Amazon Mechanical Turk (AMT) to enroll $142$ workers with diverse backgrounds to perform a user study. Each worker is required to rate a total of $26$ set of generated results. For each sample, we present the 4D avatar of different methods and corresponding text prompts. The users are required to pick the best results based on three criteria: Q1. geometry quality, Q2. appearance quality, and Q3. motion quality.

\myparagraph{Results.}
The results of evaluation metrics and user study are presented in Tab.~\ref{tab.clipscore} and Tab.~\ref{tab.user_study}, respectively. 
Tab.~\ref{tab.clipscore} demonstrates that {\name} achieves the best performance in both metrics. Compared to previous methods which separately train the canonical avatar and apply animations, our end-to-end training framework (see Algo.~\ref{alg.train}) contributes better text-3D alignment. 
With the elaborate-designed in-network motion retargeting mechanism, {\name} generates better 3D avatars while ensuring more vivid animations compared to the previous methods that posse typical optimization-by-animation paradigm. 
As shown in Tab.~\ref{tab.user_study}, the recruited evaluators further confirm that {\name}'s avatars align better with human perception. {\name} consistently achieves the highest agreement on geometry, texture, and motion quality compared to the evaluated methods.

\begin{figure*}
    \centering
          \begin{overpic}[trim=2cm 10cm 17cm 2cm,clip,width=\linewidth,grid=false]{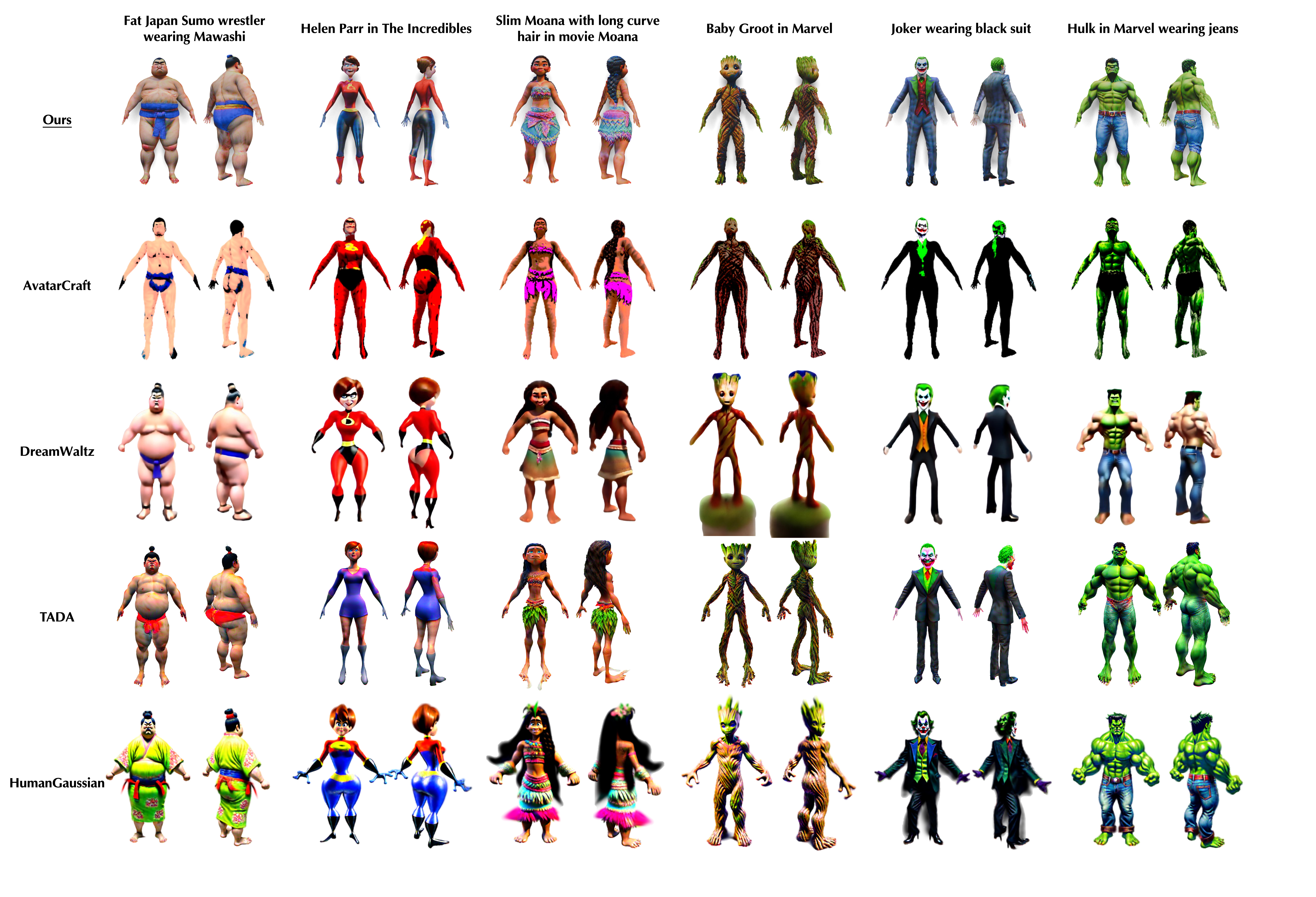}
      \end{overpic}
    \caption{\textbf{Qualitative Comparisons of the Canonical 3D Avatar.} For each method, we showcase examples of different prompts varying from real-world human to cartoon characters. The proposed {\name} exhibits better text-geometry/texture alignment while ensuring shape variance and view consistency. As result, it presents visually better results compared to prior art.}
    \label{fig.canocmp}
\end{figure*}

\begin{figure*}[ht!]
\centering
    \begin{overpic}[trim=0cm 0cm 10cm 0cm,clip,width=\linewidth,grid=false]{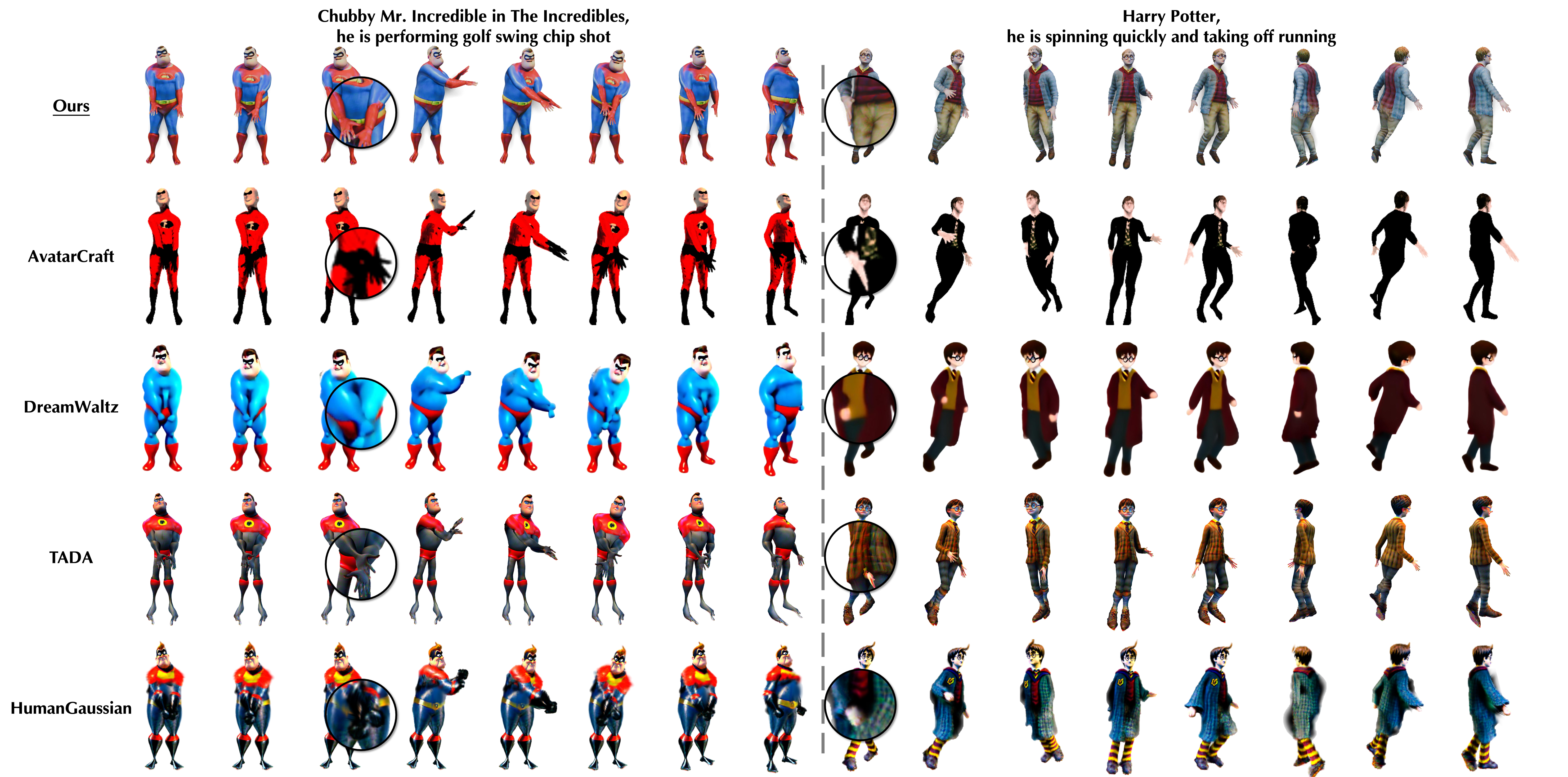}\end{overpic}
    \begin{overpic}[trim=0cm 0cm 10cm 0cm,clip,width=\linewidth,grid=false]{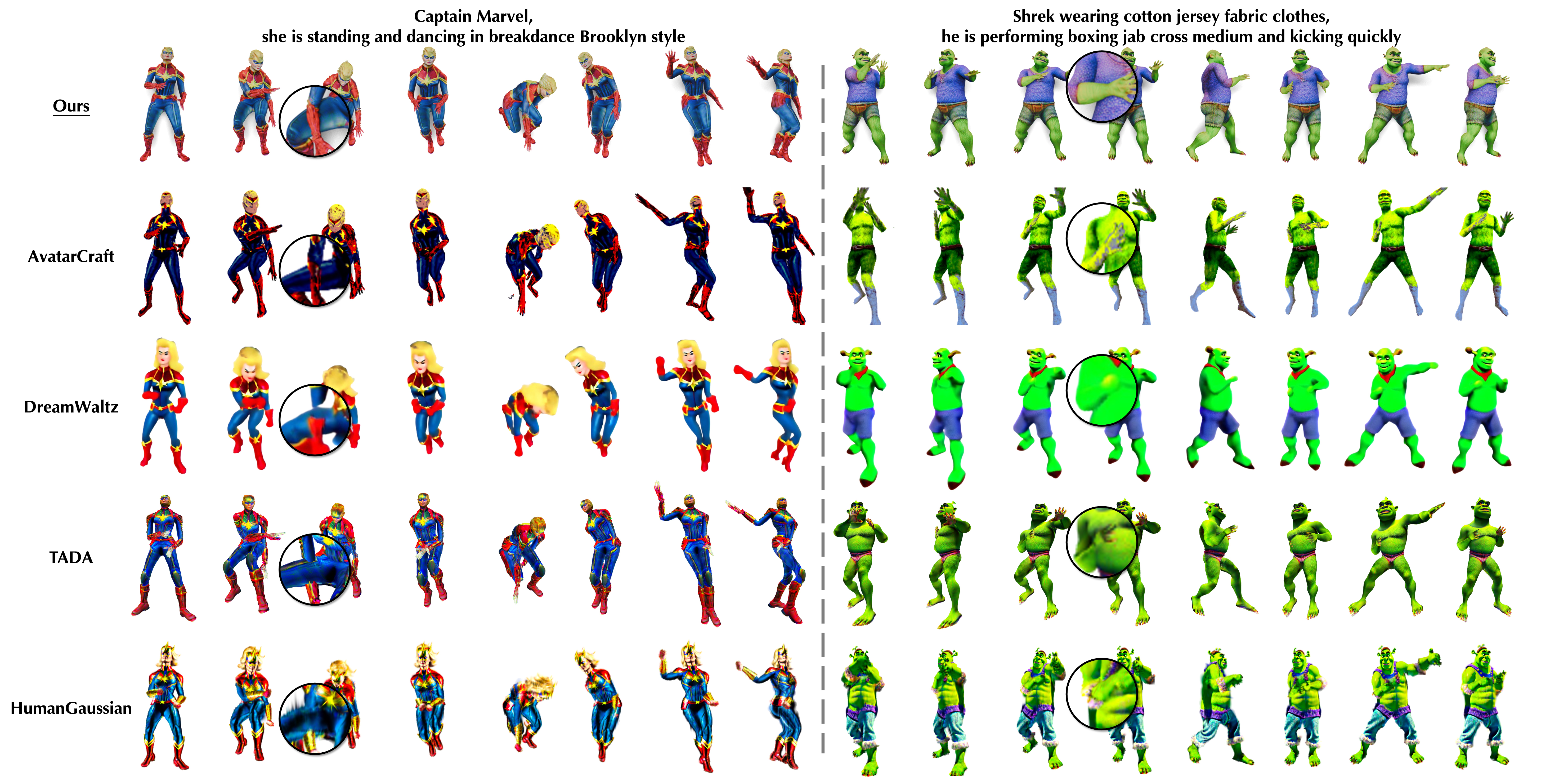}\end{overpic}
\caption{\textbf{Qualitative Comparisons of 4D Avatars.} For each sample, we showcase random selected frames of each avatar. The proposed {\name} generates vivid 4D avatars which faithfully align with the given text description. The avatars generated by the comparison methods either show deteriorated body structures or animation artifacts.}
\label{fig.animcmp}
\end{figure*}

\subsection{Qualitative Comparison}

In addition to the quantitative results, we present qualitative comparisons of the ``A-pose'' textured avatars in Fig.~\ref{fig.canocmp} with both frontal and back views.
Fig.~\ref{fig.animcmp} shows 4D avatars with randomly selected frames.
The examples in Fig.~\ref{fig.canocmp} demonstrate {\name} achieves better generation quality in terms of geometry quality, diversity and text alignment compared to the evaluated methods. Note that AvatarCraft~\cite{Jiang_2023_ICCV} uses a fixed template human body to train the NeuS~\cite{wang2021neus} and Instant-NGP~\cite{muller2022instant} representation, therefore lacks shape variance and fidelity.
Although HumanGaussian~\cite{liu2023humangaussian} and DreamWaltz~\cite{huang2023dreamwaltz} train the 3D static representation by incorporating the skeleton from canonical pose as conditions, they still suffer from Janus problem ({\eg}, Groot in different views) or view-inconsistency ({\eg}, Hulk in different views). 
TADA~\cite{liao2024tada} possesses the same 3D representation as {\name}, however, it exhibits unnatural structures of the generated avatars, due to the pose-agnostic optimization step and missing structural regularization.
As a comparison, {\name} is conceived with the crucial thinking that personalized motion plays crucial role in 4D avatar generation, with the in-optimization motions, hybrid SDS, and hierarchical regularization, {\name} generates high-fidelity 3D avatars and shows better visualized quality compared to prior arts.

Fig.~\ref{fig.animcmp} shows more samples of generated 4D avatar 8 randomly selected frames. 
Existing methods with the optimization-by-animation paradigm ignore the personalized structure of 3D avatar, which in return yields artifacts in the animated bodies ({\eg}, body piercing with arms). 
For implicit 3D representation~\cite{kerbl20233d,muller2022instant,wang2021neus}, previous methods~\cite{huang2023dreamwaltz,liu2023humangaussian,Jiang_2023_ICCV} require additional efforts to train deformable fields or build correspondence for animation but still exhibit blurry, especially with challenging poses ({\eg}, Captain Marvel's examples). 
For explicit 3D representation~\cite{pavlakos2019expressive}, while such mesh-based representation is flexible to animate with arbitrary motions without the requirements of fine-tuning, the animated avatars of TADA~\cite{liao2024tada} are still unsatisfactory since the canonical avatar do not preserve reasonable body structures. 
Simply copying motions introduces additional artifacts.
As a comparison, {\name} produces 4D avatars end-to-end, distinguishing the shape-level and skeleton-level differences through motion retargeting.
This allow {\name} to generate reasonable 3D avatars while possessing accurate and vivid 4D animations.

\subsection{Ablation Studies} \label{sec.abl}

\begin{figure*}
    \centering
    \subfloat[A1: w/o hybrid SDS]{
    \begin{overpic}[trim=0cm 0cm 0cm 0cm,clip,width=0.287\linewidth,grid=false]{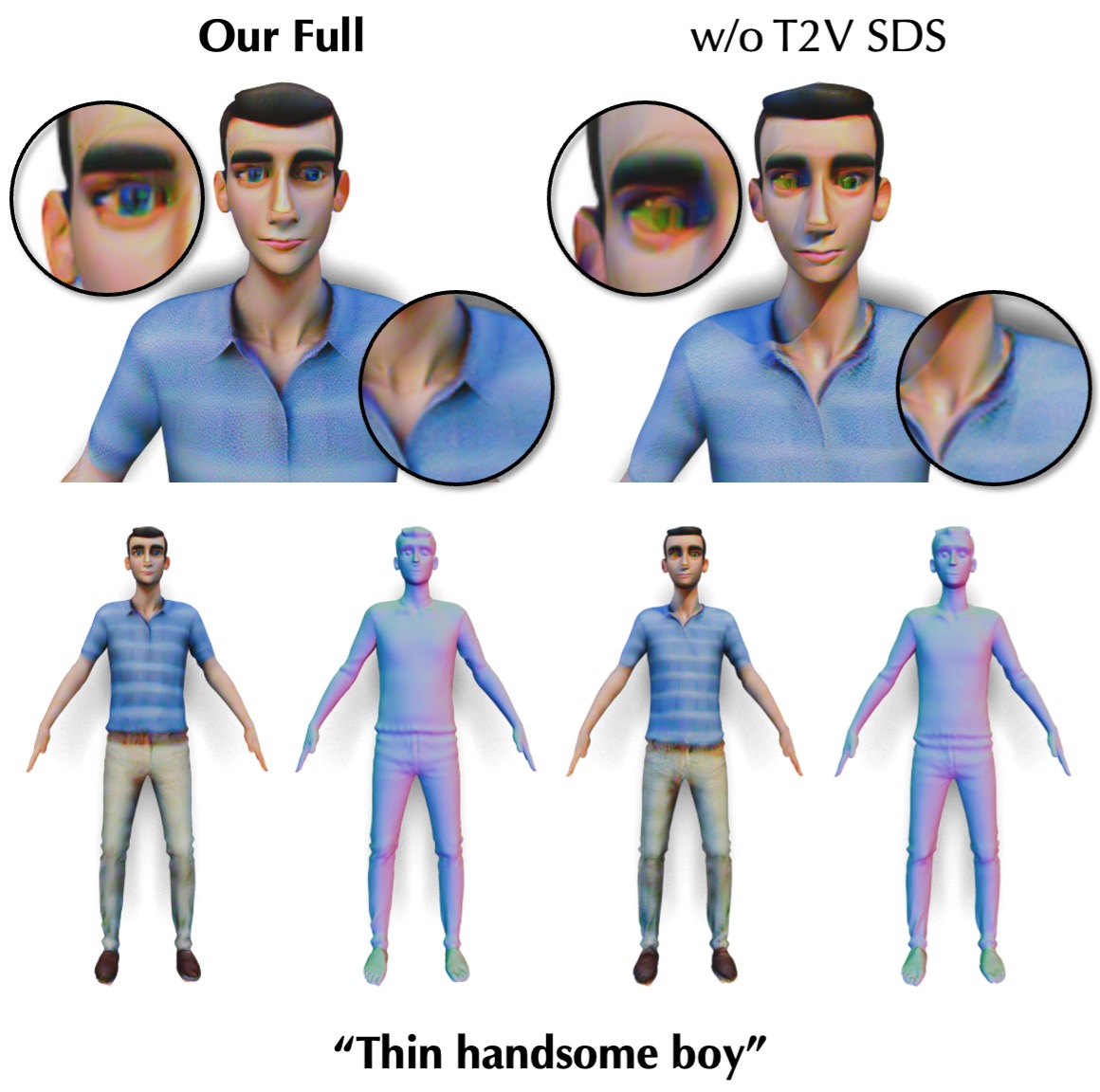}\end{overpic}
    \label{fig.abl.wot2v}
    }
    \subfloat[A2: w/o revised SDS]{
    \begin{overpic}[trim=0cm 0cm 0cm 0cm,clip,width=0.38879\linewidth,grid=false]{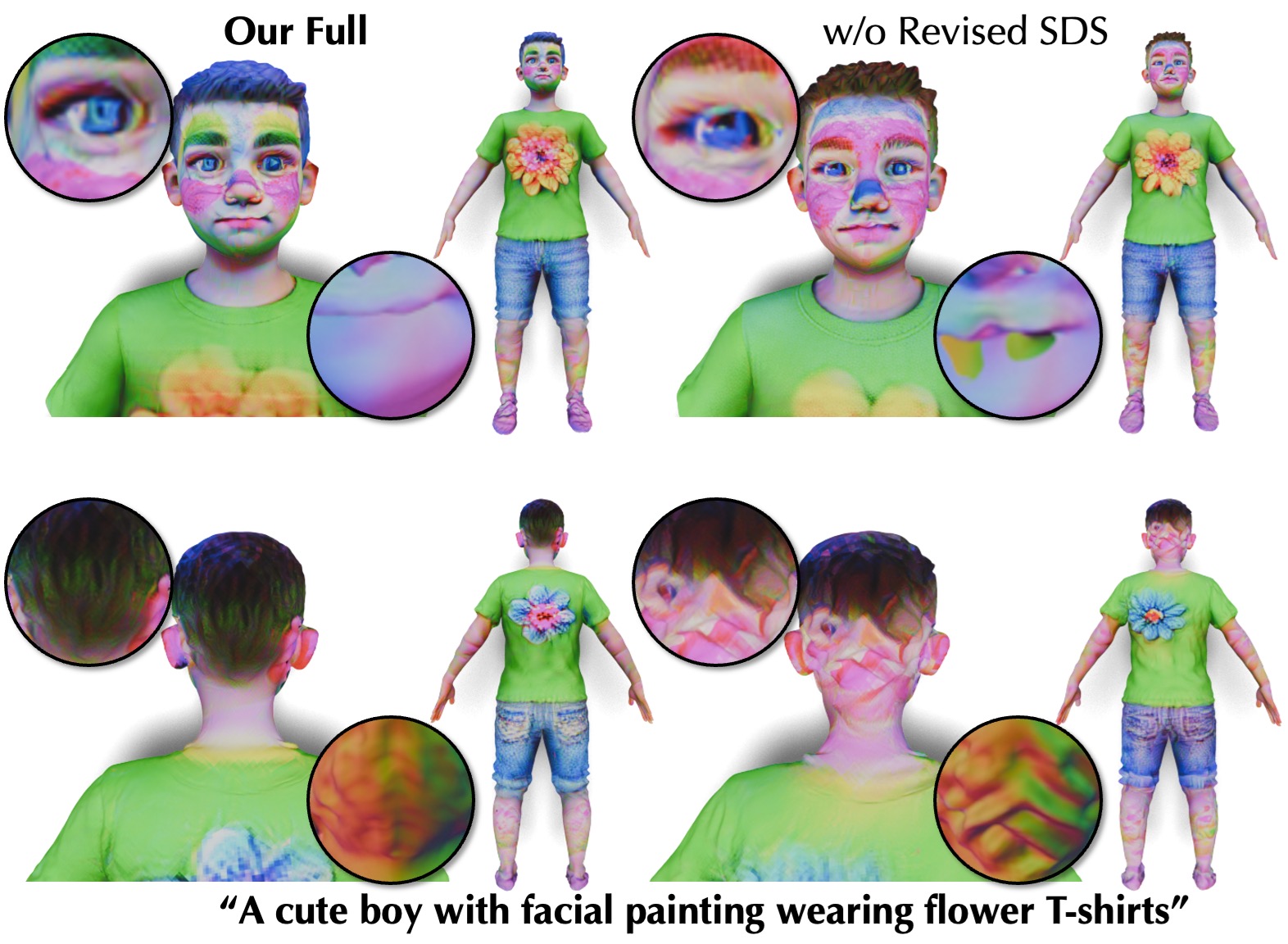}\end{overpic}
    \label{fig.abl.womask}
    }  
    \subfloat[A3: w/o pose sampling]{
    \begin{overpic}[trim=0cm 0cm 0cm 0cm,clip,width=0.27407595\linewidth,grid=false]{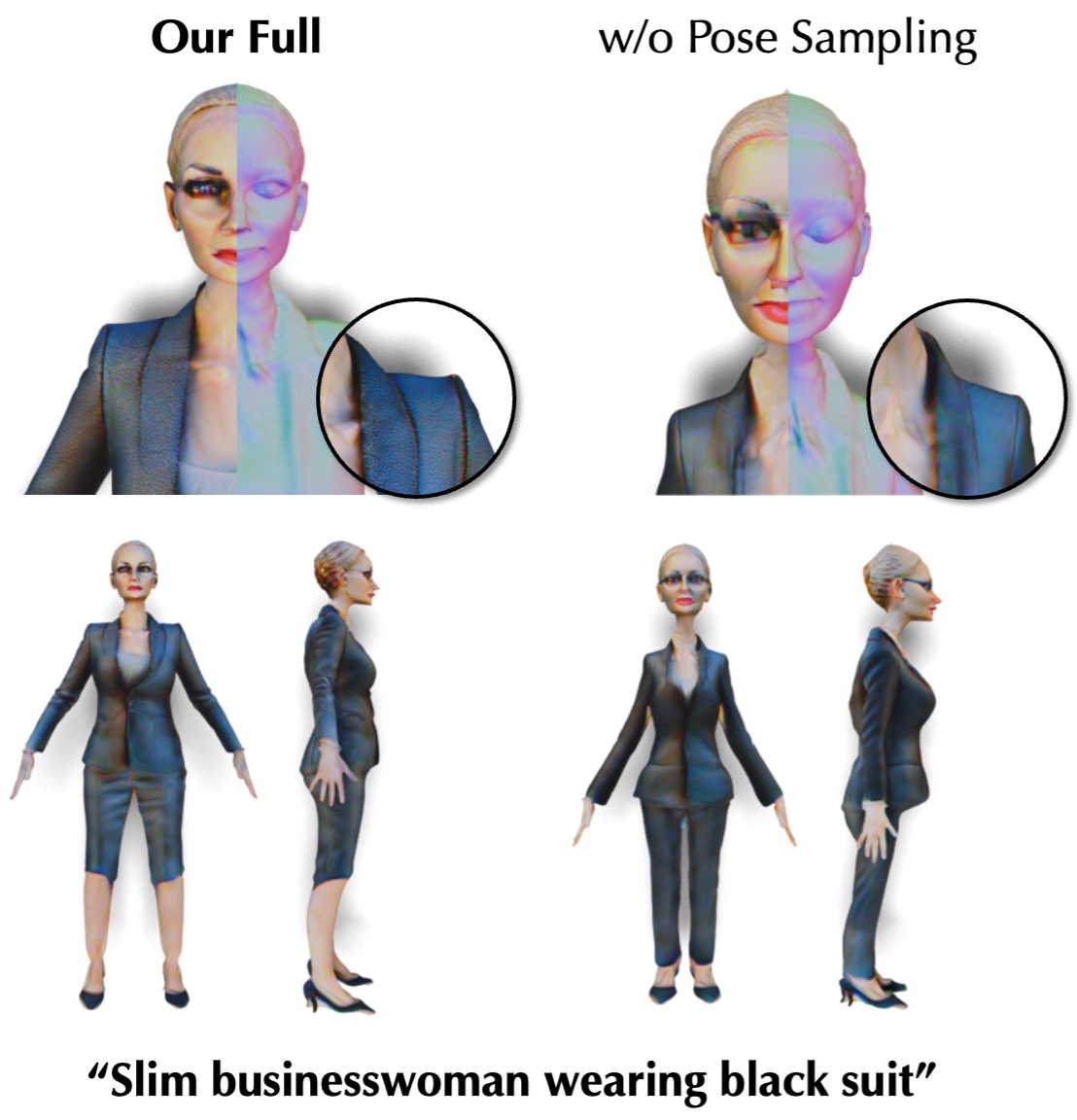}\end{overpic}
    \label{fig.abl.wopose}
    }  

    \subfloat[A4: w/o in-network motion retargeting]{
    \begin{overpic}[trim=0cm 0cm 0cm 0cm,clip,clip,width=0.49592\linewidth,grid=false]{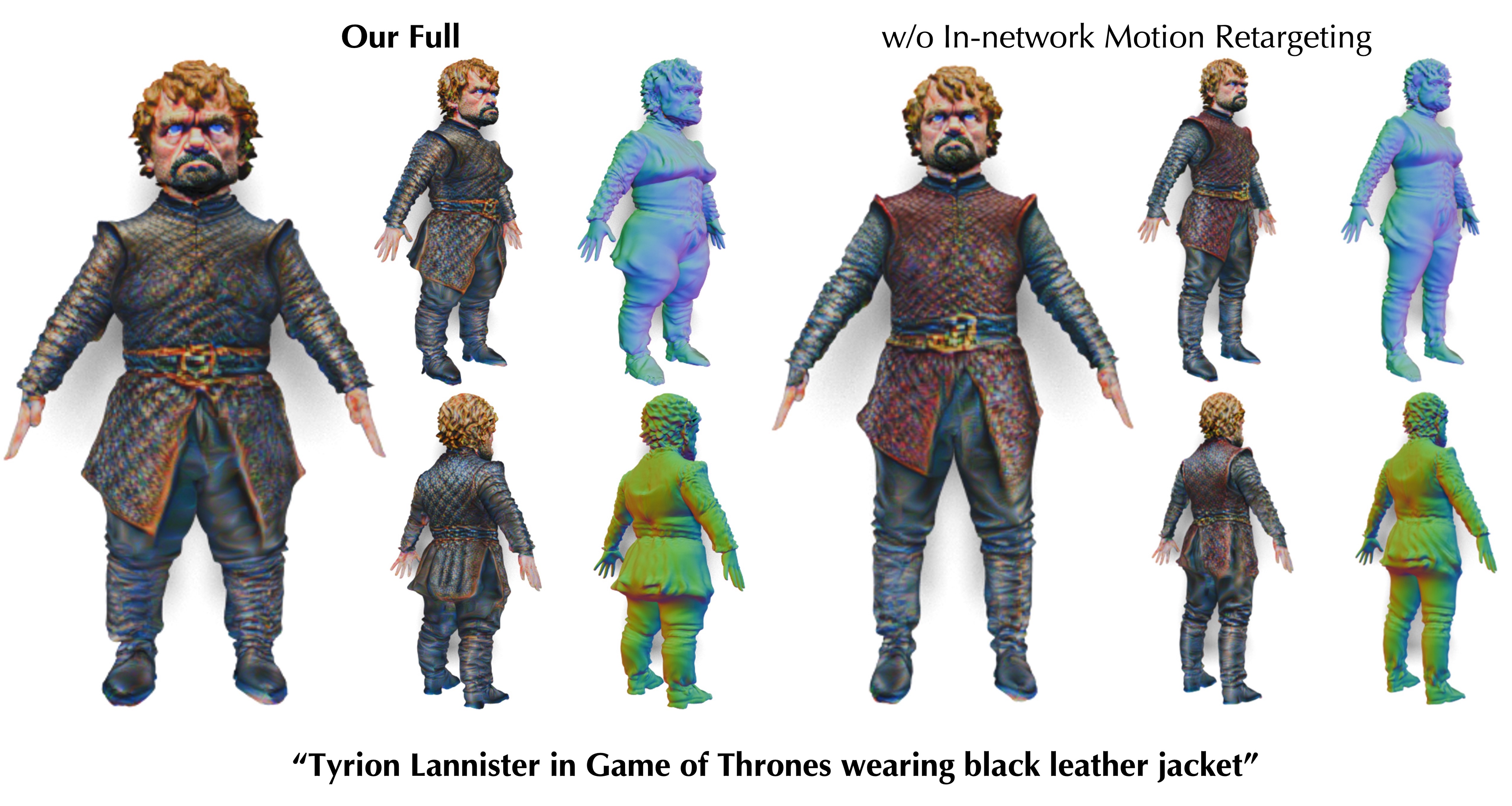}\end{overpic}
    \label{fig.abl.woretarget}
    }
    \subfloat[A5: w/o hierarchical regularization]{
    \begin{overpic}[trim=0cm 0cm 0cm 0cm,clip,clip,width=0.45407\linewidth,grid=false]{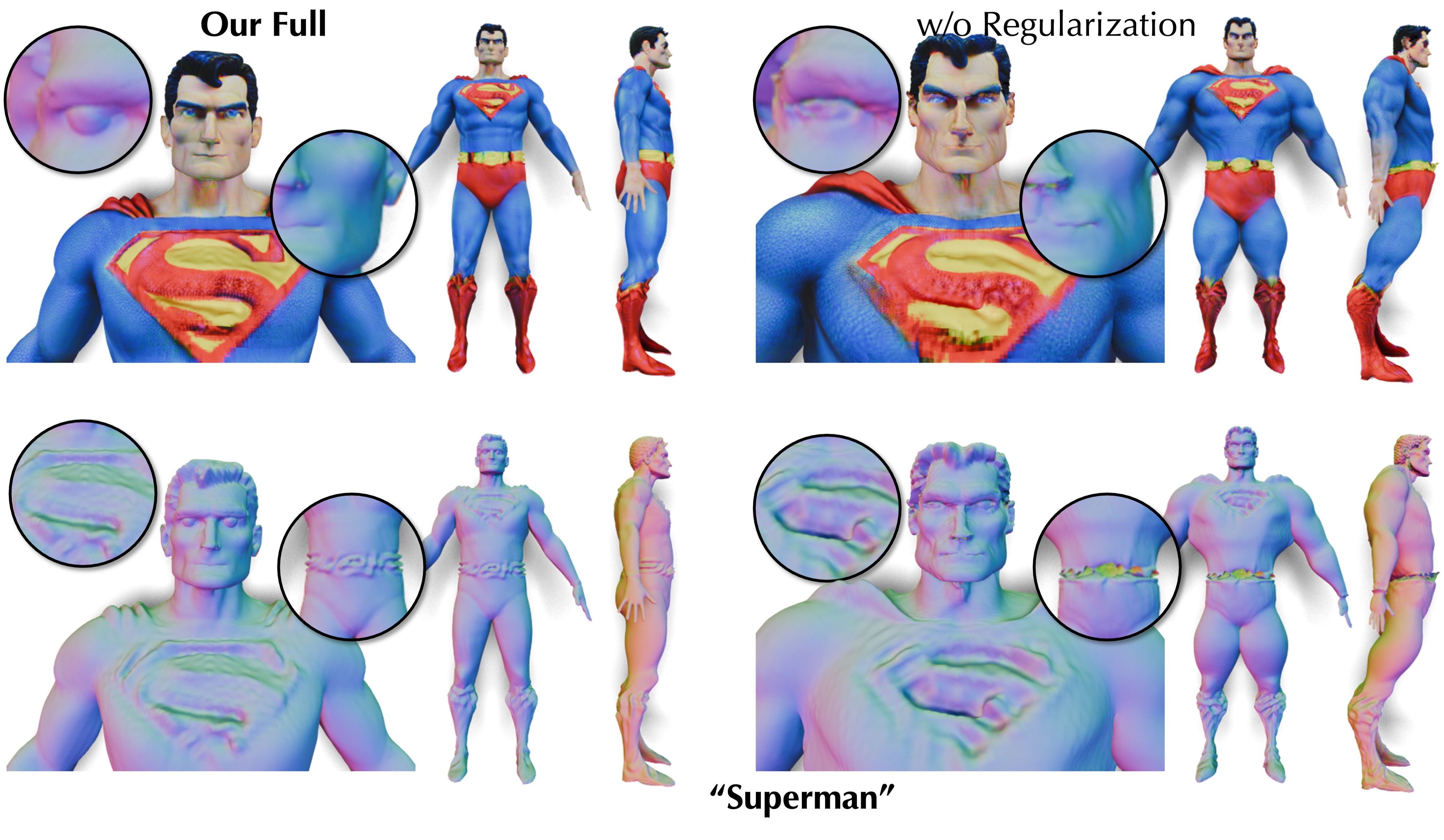}\end{overpic}
    \label{fig.abl.woreg}
    }
    \caption{\textbf{Ablation Study of {\name}}. We qualitatively demonstrate the necessity of each component in {\name}. For each ablation, we showcase: \textit{Left}. Generated 3D avatar from {\name}, \textit{Right}. Results after removing the stated component.}
    \label{fig.abl}
\end{figure*}

We conduct detailed ablation studies to demonstrate the effectiveness of each component in {\name}. We subsequently compare the results without (w/o) the following components: \textbf{A1}. w/o hybrid SDS, instead with only T2I-based SDS ({\ie}, Eq.~\ref{eq.skelsds}); \textbf{A2}. w/o skeleton occlusion detection and facial mask for our revised SDS, instead, simply stack T2V and T2I models with skeleton-conditioned SDS; \textbf{A3}. w/o pose sampling and motions, instead with only canonical pose and T2I-based SDS; \textbf{A4}. w/o in-network motion retargeting, instead with fixed initialized motions; \textbf{A5}. w/o hierarchical regularization. 
In each ablation, we preserve other parts unchanged and remove only necessary components for comparison. The qualitative results are shown in Fig.~\ref{fig.abl}.

Fig.~\ref{fig.abl}\subref{fig.abl.wot2v} shows the necessity of our hybrid SDS. T2V prior provides view-consistent supervision and contributes to textural alignment and fidelity. Without T2V supervision, the generated avatar ({\eg}, ``Thin handsome boy'') exhibits unnatural textures ({\eg}, eyes) and lack fine details ({\eg}, collars).
However, Fig.~\ref{fig.abl}\subref{fig.abl.womask} demonstrates that simply stacking T2V and T2I is harmful to avatar fidelity, without the elaborate-designed skeleton-conditioned SDS (Eq.~\ref{eq.skelsds}) and masked sequential SDS (Eq.~\ref{eq.seqsds}), the generated avatars exhibit Janus problems and less expressiveness of facial appearance ({\eg}, confusion of head for ``A cute boy with facial painting''). 

Intuitively, while directly adding motion retargeting can address the artifacts for pose-hoc animation, our ablation study shows the importance of in-network motion retargeting in optimization stage.
First, Fig.~\ref{fig.abl}\subref{fig.abl.wopose} shows that optimization with a fixed canonical pose is harmful to the body structure. As we discussed before, the lack of pose diversity yields potential distribution bias of the pretrained diffusion models. This makes the SDS-based optimization learn flawed geometry and fail to capture reasonable body scale ({\eg}, disproportionate head and limbs of ``Slim businesswoman'').
In addition, Fig.~\ref{fig.abl}\subref{fig.abl.woretarget} shows comparisons of simply sampling the initial motion to improve the pose diversity for optimization, as attempted by previous methods~\cite{kolotouros2023dreamhuman,yuan2023gavatar,zhang2023avatarstudio}. However, such intuitive sampling ignores personalized skeleton and geometry, making the 3D representation learns a template-like geometry that compromises the mismatched motions ({\eg}, ``Tyrion Lannister'' in fact is with \textit{Dwarfism}). In contrast, {\name} progressively updates mesh and motion to achieve diverse 3D avatar with vivid animations, which better aligns with the given description.

Finally, Fig.~\ref{fig.abl}\subref{fig.abl.woreg} demonstrates the effectiveness of our simple-yet-effective hierarchical regularization, where the shape constraint imposes the avatar to follow a reasonable body structure. For example, the Laplacian smoothness loss contributes to fine details ({\eg}, belts of ``Superman''), and the facial regularization improves the misaligned lips and intersected eyes.
\section{Limitations \& Future Work}

Section~\ref{sec:experiment} has shown that {\name} can produce high-quality 4D avatar. However, there are still several limitations. For example, as we use SMPL-X~\cite{pavlakos2019expressive} as 3D human prior, {\name} cannot represent out-of-distribution text descriptions such as animals, and the per-vertex displacement is hard to represent hair strands, eyelashes, and complex accessories. We leave it for future work to integrate implicit representation for learning more complex avatars.

\section{Conclusion}

In this paper, we propose {\name} to generate high-quality avatars with vivid animations from only text descriptions. {\name} is conceived by the crucial thinking that motion diversity and fidelity play crucial roles in text-based 4D avatar generation. We consider the skeleton and geometry difference between the source template and the target avatar, by proposing in-network motion retargeting to progressively optimize geometry, texture, and motion. We also propose a novel hybrid occlusion-aware SDS by leveraging the power of skeleton-conditioned T2I and T2V models, with simple-yet-effective hierarchical regularization to achieve high-fidelity 4D avatar generation quality. Both quantitative and qualitative experiments demonstrate {\name} can generate vivid 4D avatar and outperforms previous methods. In-depth ablations show the effectiveness of each component in {\name}.

\bibliographystyle{IEEEtran}
\bibliography{main}

\end{document}